\documentclass{article}

\usepackage[preprint]{epiguard}

\usepackage{amsmath,amsfonts,bm}

\def\eqref#1{equation~\ref{#1}}

\def\1{\bm{1}}

\DeclareMathAlphabet{\mathsfit}{\encodingdefault}{\sfdefault}{m}{sl}
\SetMathAlphabet{\mathsfit}{bold}{\encodingdefault}{\sfdefault}{bx}{n}

\usepackage{graphicx}
\usepackage{booktabs}
\usepackage{wrapfig}
\usepackage{caption}
\usepackage{array}
\usepackage{hyperref}
\usepackage{placeins}

\definecolor{EpiBlue}{RGB}{14,62,120}
\hypersetup{
  colorlinks=true,
  linkcolor=black,
  citecolor=blue,
  urlcolor=EpiBlue
}

\newcommand{\systemname}{\textsc{EpiGuard}}

\title{%
EpiGuard: An Autonomous Infectious-Disease Early Warning\\ Agent Platform
}

\author{%
  \makebox[\dimexpr\textwidth-6pt\relax][c]{%
    Tonglong Wei\textsuperscript{*}\quad
    Zheng Yuan\textsuperscript{*}\quad
    Lingshuai Lin\quad
    Yuhui Yao\quad
    Xing Xie\quad
    Na Zou\textsuperscript{\ensuremath{\dagger}}\quad%
 }\\[0.35em]
  \makebox[\dimexpr\textwidth-6pt\relax][c]{%
    Shanghai Artificial Intelligence Laboratory
  }\\[0.35em]
  \makebox[\dimexpr\textwidth-6pt\relax][c]{%
    \normalfont\footnotesize
    \textsuperscript{*}Equal contribution.\quad
    \textsuperscript{\ensuremath{\dagger}}Corresponding author: \texttt{zouna@pjlab.org.cn}.%
  }\\[0.4em]
  \makebox[\dimexpr\textwidth-6pt\relax][c]{%
    \href{https://epiguard.intern-ai.org.cn/}{%
      \textsf{\bfseries\textcolor{EpiBlue}{https://epiguard.intern-ai.org.cn/}}%
    }%
  }
}

\date{}
\begin{document}

\maketitle

\begin{abstract}
Infectious disease outbreaks often spread rapidly before signs of widespread transmission become apparent, leaving a big challenge for rapid response and effective intervention.
However, current infectious-disease early-warning systems face three major challenges: fragmented epidemiological evidence across diverse surveillance sources, highly uneven data availability across diseases and geographic regions, and delayed warning procedure caused by manual data collection, analysis, and reporting.
To address these challenges, we present \systemname, an autonomous, continually self-improving agentic platform for timely and reliable early warning of more than 50 infectious diseases across different regions.
\systemname\ proactively monitors and captures newly released official surveillance records, using the latest available evidence for risk assessments. 
It dynamically adjusts its analytical strategy across three evidence availability regimes, enabling reliable risk assessment under diverse data conditions, ranging from the absence of historical observations to sparse and abundant evidence. 
Moreover, \systemname\ learns from the successful execution process by preserving effective data-acquisition and analytical procedures as reusable standard operating procedures, enabling it to progressively generalize to previously unseen diseases and regions.
We use officially released infectious-disease surveillance data from the China CDC and the United States CDC to evaluate \systemname’s effectiveness.
Experimental results demonstrate that \systemname\ outperforms state-of-the-art forecasting systems, improving forecasting performance by 9.68\% and 14.04\% on the China CDC and U.S. CDC datasets, respectively.

\end{abstract}


\setcounter{section}{0}
\section{Introduction}
\label{sec:introduction}

Infectious-disease outbreaks spread rapidly and affect large populations in a short time. The World Health Organization emphasizes early detection, verification, and rapid response as key components of public-health surveillance~\cite{world2023early}. 
Timely and reliable early warning supports these goals by identifying infectious disease signals and providing relevant information for risk assessment and response.

To enable effective infectious-disease early warning, surveillance systems have evolved from statutory case reporting~\cite{jajosky2004evaluation}, syndromic surveillance~\cite{hughes2020emergency}, and statistical threshold-based detection~\cite{farrington1996statistical} to internet-based event surveillance~\cite{keller2009unstructured} and multi-source epidemiological modeling~\cite{santillana2015combining,meckawy2022effectiveness}. For example, the China Infectious Disease Automated Alert and Response System~\cite{yang2011nationwide} generates warning signals from statutory case reports using fixed thresholds and time-series models, followed by local verification and epidemiological investigation. In parallel, digital epidemic-intelligence systems such as HealthMap~\cite{freifeld2008healthmap}, GPHIN~\cite{dion2015big}, and ProMED-mail~\cite{carrion2017promed} monitor news reports, official announcements, and other open-source information to identify emerging infectious disease signals~\cite{brownstein2008sans}. By incorporating multiple public information, such approaches can support earlier detection and more reliable infectious-disease warning.

Despite these advances, existing infectious-disease early-warning systems remain constrained by three fundamental challenges. 
\textbf{First, epidemiological evidence is fragmented across diverse surveillance sources.} Infectious-disease surveillance draws on several types of evidence, ranging from case and laboratory records to official and open-source reports. These sources are maintained separately and updated at different times. This fragmentation makes it difficult to bring together complete and up-to-date evidence in time to support accurate early warnings.
\textbf{Second, evidence availability is highly uneven across diseases and geographic regions.} Some diseases are routinely monitored and have years of historical records. By contrast, emerging diseases and regions with limited surveillance may have only a few reports and little or no historical data. A single forecasting method may therefore not work equally well across all diseases and regions.
\textbf{Third, delayed warning procedure caused by manual data collection, analysis, and reporting.} To forecast disease trends and generate early warnings, epidemiologists need to collect, clean, and analyze data from multiple sources. This labor-intensive process introduces substantial delays between the emergence of epidemic signals and warning delivery, limiting the ability of existing systems to respond rapidly.

\begin{figure}[t]
\centering
\includegraphics[width=0.8\linewidth]{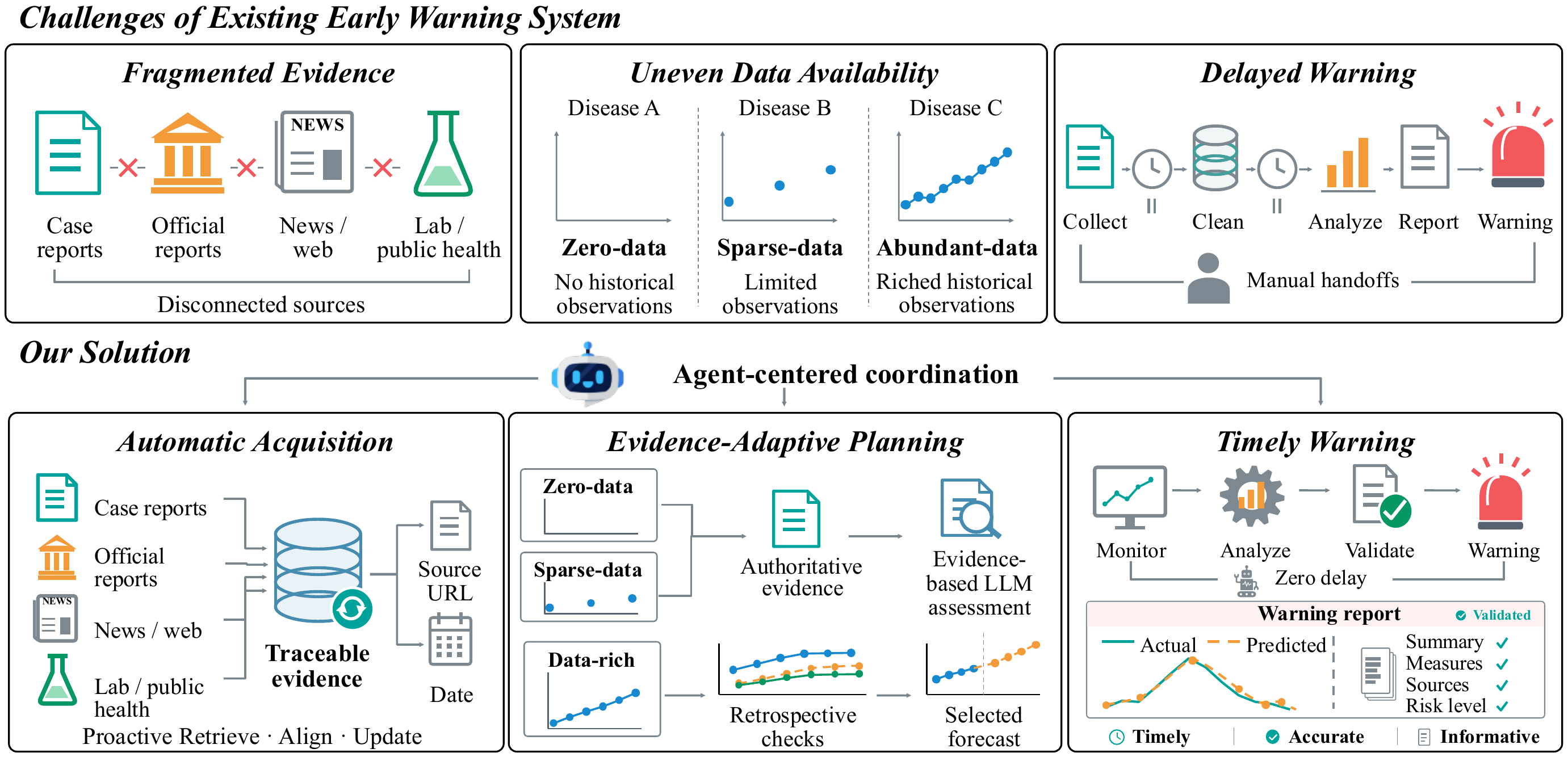}
\caption{Limitations of infectious disease early warning systems and our solution.}
\label{fig:challenges}
\end{figure}

To address these challenges, we present \systemname, an autonomous and continually self-improving agentic platform for timely and reliable infectious-disease early warning, as shown in Figure~\ref{fig:challenges}.
\systemname\ maintains an up-to-date epidemiological evidence repository by monitoring diverse surveillance sources, capturing newly released records, and merging them with previously acquired evidence. To address uneven evidence availability, the platform employs an adaptive planning mechanism that combines LLM reasoning with a forecasting model pool to dynamically adapt the analytical workflow across three evidence regimes---absent, sparse, and abundant. 
Finally, \systemname\ incorporates a self-improvement mechanism that converts effective data-acquisition and analysis procedures from successful warning tasks into reusable standard operating procedures (SOPs). These SOPs are progressively refined through subsequent tasks, supporting the extension to new diseases and regions.

We use officially released infectious-disease surveillance data from the China CDC and the U.S. CDC to evaluate \systemname's effectiveness. Experimental results show that \systemname\ consistently outperforms the strongest baselines across diseases and regions, achieving average improvements of 9.68\% on the China CDC data and 14.04\% on the U.S. CDC data. 

The main contributions of this work are summarized as follows:

\begin{itemize}
    \item We present \systemname, an autonomous and self-improving agentic platform that unifies evidence acquisition, analysis planning, forecasting, risk assessment, and report generation for timely and reliable infectious-disease early warning of more than 50 infectious diseases across different regions.

    \item We introduce an evidence-adaptive planning mechanism that combines LLM reasoning with a forecasting model pool and dynamically configures analytical workflows across absent, sparse, and abundant evidence regimes. This mechanism enables task-adaptive analysis without relying on a fixed pipeline or a single forecasting model.

    \item We develop a self-improvement mechanism that preserves effective data acquisition strategies and analysis workflows from previous tasks. These procedural capabilities can be retrieved and refined in subsequent tasks, allowing accumulated experience to support new diseases and regions.

    \item We use officially released infectious disease data from the China CDC and U.S. CDC.  Experimental results show that \systemname\ consistently outperforms strong baselines across diseases and regions.
\end{itemize}




\section{Platform}\label{sec:method}

\subsection{System architecture}

\begin{figure}[t]
\centering
\includegraphics[width=\linewidth]{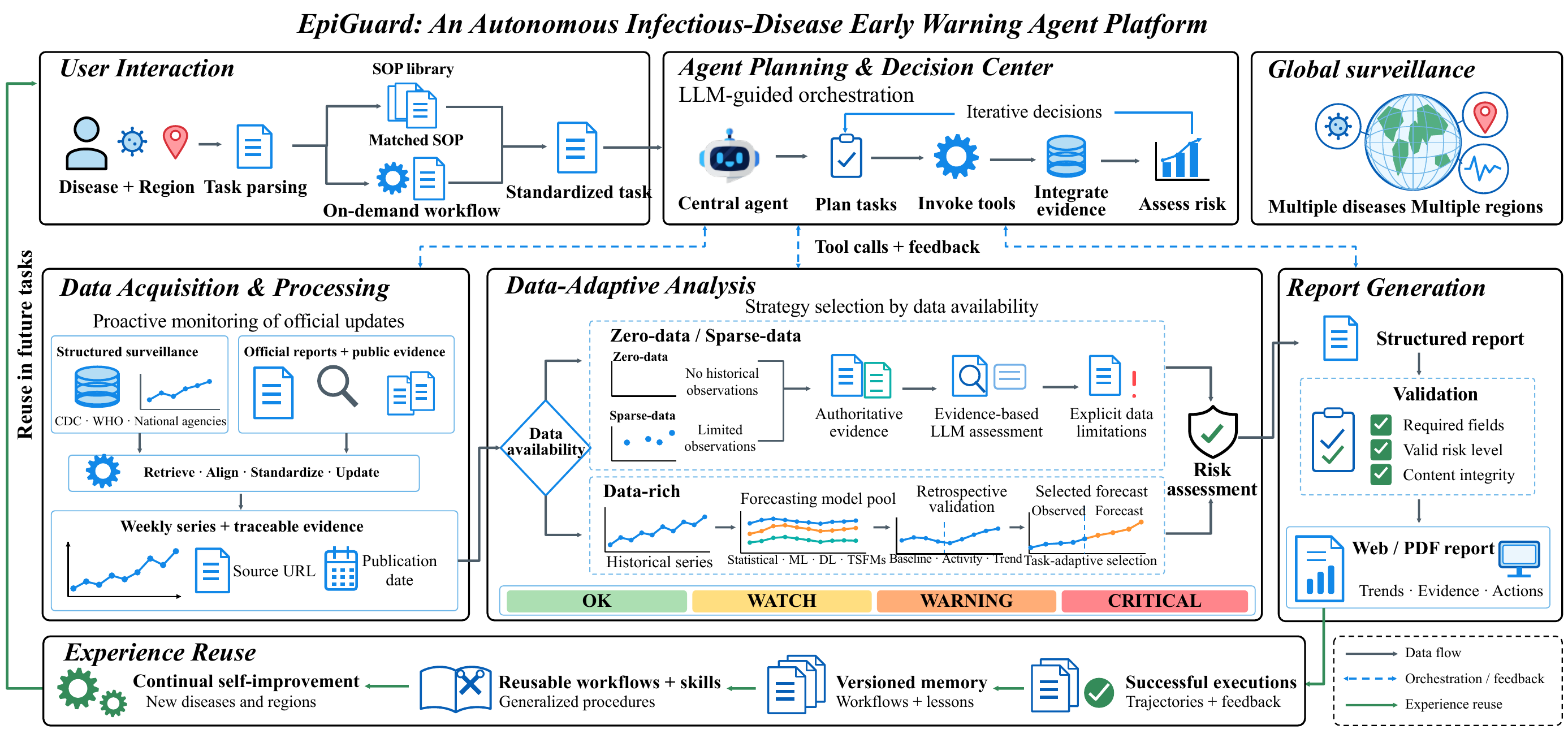}
\caption{The framework of EpiGuard.}
\label{fig:framework}
\end{figure}

The overall architecture of \systemname\ is illustrated in Figure~\ref{fig:framework}. At its core is the \textbf{Agent Planning and Decision Center (APDC)}, which coordinates five functional components: the User Interaction Layer (UIL), Data Acquisition and Processing (DAP), Model Analysis and Risk Determination (MARD), Report Generation (RG), and the Self-improvement Mechanism. These components form a closed-loop early-warning framework that connects user requests, epidemiological evidence acquisition, evidence-adaptive analysis, risk assessment, report generation, and experience accumulation.

The \textbf{UIL} serves as the interface between open-ended user requests and \systemname's analytical capabilities. It extracts the target disease and geographic region from each request. The resulting disease--region pair is then used to retrieve the corresponding standard operating procedure (SOP).
Guided by the SOP, the \textbf{APDC} performs adaptive planning by dynamically selecting appropriate analytical strategies, forecasting models, and tools based on evidence availability. It coordinates the corresponding tool calls across the SOP-defined stages of data acquisition, evidence analysis, forecasting, risk determination, and report generation.
The \textbf{DAP} component maintains an up-to-date epidemiological evidence repository by monitoring diverse surveillance sources and integrating newly released records with previously acquired evidence.
The \textbf{MARD} component dynamically adapts the analytical workflow across three evidence regimes---absent, sparse, and abundant. It coordinates LLM reasoning with a forecasting model pool to determine risk according to the available evidence.
The \textbf{RG} component organizes the collected evidence, analytical results, and risk assessment into a regime-specific report. It further performs schema and content checks before rendering the report as a web or PDF artifact.
The \textbf{Self-improvement Mechanism} captures effective data-acquisition and analysis procedures from successful warning tasks as reusable SOPs, which can be retrieved to guide subsequent tasks.

\subsection{User Interaction Layer}

Early-warning requests may concern different diseases and geographic regions, while the available data sources and standard operating procedures (SOP) vary across disease--region pairs. The User Interaction Layer (UIL) connects these requests to \systemname's analytical capabilities by identifying the target disease $d$ and geographic region $r$ and mapping them to a corresponding SOP.

To organize the available SOPs, \systemname\ maintains a SOP set: $ \mathcal{G} = \left\{ \left(r_i,\mathcal{D}_i,\mathcal{S}_i\right) \right\}_{i=1}^{N}$,
where $N$ is the number of geographic regions and disease pairs, $\mathcal{S}$ is the corresponding SOP set, which defines the surveillance data route and reusable operations for data acquisition, preprocessing, analysis, and report generation. By indexing SOPs according to disease--region pairs, $\mathcal{G}$ enables the UIL to retrieve a corresponding SOP when available and identify requests that must be routed to the zero-shot workflow otherwise.


\subsection{Agent Planning and Decision Center}

The Agent Planning and Decision Center (APDC) is the LLM-based planning and orchestration component of \systemname. It receives the request information $q=(r,d)$ and the SOP retrieval result $s^{\star}$ from the UIL, where $s^{\star}=\varnothing$ indicates that no dedicated SOP is available. Conditioned on these inputs, the APDC uses an LLM's reasoning capability to decompose the request into executable actions and organize them into an adaptive analysis plan. It subsequently coordinates DAP, MARD, and RG by determining their invocation order and routing intermediate results among them.

Let $\mathcal{T}=\{\mathcal{T}_{\mathrm{DAP}},\mathcal{T}_{\mathrm{MARD}},\mathcal{T}_{\mathrm{RG}}\}$ denote the tools exposed by the three operational components. The APDC constructs the execution plan as
\begin{equation}
\mathcal{P}(q,s^{\star})
=
\Phi_{\mathrm{LLM}}
\left(q,s^{\star},\mathcal{T}\right)
=
\left(
a_{\mathrm{DAP}},
\left\{\mathcal{P}_{z}\right\}_{z\in\mathcal{Z}},
a_{\mathrm{RG}}
\right),
\end{equation}
where $\Phi_{\mathrm{LLM}}$ denotes LLM-based planning, $a_{\mathrm{DAP}}$ represents the actions required for evidence acquisition and processing, and $a_{\mathrm{RG}}$ denotes report-generation actions. The set $\mathcal{Z}= \left\{ \mathrm{absent}, \mathrm{sparse}, \mathrm{abundant} \right\}$ contains the three evidence-availability regimes, with $\mathcal{P}_{z}$ denoting the corresponding analysis branch executed through MARD. This adaptive planning mechanism dynamically adjusts the analytical workflow by selecting appropriate evidence regimes according to the availability of the evidence.

During execution, the APDC first invokes DAP to obtain the available epidemiological evidence. When $s^{\star}$ is available, the LLM uses the retrieved SOP as procedural context for planning; otherwise, it follows the zero-shot workflow based on publicly available epidemiological information. After DAP returns the collected evidence, the APDC determines its availability regime $z$ and activates the corresponding branch $\mathcal{P}_{z}$. The selected branch is dispatched to MARD for evidence analysis, forecasting, and risk determination, after which MARD forwards the outputs to RG for report construction. 
In this process, the LLM within the APDC is responsible for planning and coordination, while specialized tools perform the concrete analytical operations for each task.

\subsection{Data Acquisition and Processing}

To support timely early warning, the Data Acquisition and Processing (DAP) maintains an up-to-date epidemiological evidence repository. DAP proactively monitors configured surveillance sources, detects newly released records, and integrates these changes with previously acquired evidence. It further provides a unified processing interface for surveillance systems that differ in geographic region, reporting frequency, data format, and epidemiological indicators.

\paragraph{Monitored surveillance sources.}
Table~\ref{tab:surveillance-sources} summarizes the official surveillance sources currently monitored by \systemname. These sources provide different epidemiological measurements, including reported case counts, laboratory-confirmed detections, and testing positivity. Each structured source is associated with a source-specific acquisition procedure referenced by its SOP. The procedure specifies how DAP accesses, parses, and processes records from that source, allowing the system to monitor diverse surveillance systems through a consistent interface.

\begin{table}[t]
\centering
\small
\caption{Official surveillance sources currently integrated into \systemname.}
\label{tab:surveillance-sources}
\begin{tabular}{p{0.21\linewidth}p{0.14\linewidth}p{0.18\linewidth}p{0.37\linewidth}}
\toprule
\textbf{Source} & \textbf{Coverage} & \textbf{Frequency} & \textbf{Surveillance content} \\
\midrule
China CDC Weekly~\cite{chinacdcRespiratory} & China & Weekly & Infectious-disease surveillance reports and epidemiological updates on public-health events. \\
\hline

WHO FluNet~\cite{who_flunet} & Global & Weekly & Laboratory-confirmed influenza detections by country, virus type, and subtype. \\
\hline
U.S. CDC NNDSS~\cite{cdcNNDSS} & United States & Weekly & Provisional case notifications for nationally notifiable diseases reported by states and territories. \\
\hline

UKHSA Data Dashboard~\cite{ukhsa_data_dashboard} & England & Regularly updated & Case counts, testing positivity, respiratory-virus activity, and other infectious-disease indicators. \\
\hline

ECDC Surveillance Atlas / ERVISS~\cite{ecdc_surveillance_atlas} & EU/EEA & Weekly & Surveillance data for multiple infectious diseases and weekly monitoring of influenza, RSV, and SARS-CoV-2. \\
\hline

PHAC FluWatch+~\cite{phac_fluwatch_plus} & Canada & Weekly & National indicators for influenza, COVID-19, RSV, and other respiratory viruses. \\
\hline

Singapore Weekly Infectious Diseases Bulletin~\cite{sgCDA} & Singapore & Weekly & Current and cumulative notifiable-disease counts, historical comparisons, and selected respiratory indicators. \\
\hline

Japan Infectious Diseases Weekly Report~\cite{jihsIDWR} & Japan & Weekly & Reported cases, temporal trends, and historical comparisons for designated infectious diseases. \\
\bottomrule
\end{tabular}
\end{table}

\paragraph{Proactive monitoring and repository update.}
DAP monitors each configured source according to its reporting schedule rather than waiting for a user request to initiate data acquisition. At each monitoring cycle, it retrieves the most recent release and compares the returned records with the corresponding observations stored in the evidence repository. Let $\mathcal{R}_{t-1}$ denote the stored records before the current monitoring cycle and $\mathcal{D}_{t}$ denote the records obtained from the source. The repository is updated as
\begin{equation}
\Delta_t
=
\operatorname{Compare}
\left(
\mathcal{D}_{t},
\mathcal{R}_{t-1}
\right),
\qquad
\mathcal{R}_{t}
=
\operatorname{Merge}
\left(
\mathcal{R}_{t-1},
\Delta_t
\right),
\end{equation}
where $\Delta_t$ contains newly reported observations and revisions to previously released records. DAP records the retrieval time, most recent reporting period, and source provenance as repository metadata. This monitoring process decouples evidence maintenance from individual requests and prevents downstream analyses from relying on manually prepared or outdated data snapshots.


\paragraph{Unified evidence representation.}
For a structured surveillance indicator $m$, DAP represents the observations associated with disease $d$ in region $r$ as
\begin{equation}
\mathbf{x}_{r,d}^{(m)}
=
\left(
x_{r,d}^{1},
x_{r,d}^{2},
\ldots,
x_{r,d}^{T}
\right),
\end{equation}
where $T$ is the number of reporting periods and $x_{r,d}^{t}$ is the observed value during period $t$. DAP returns the acquired evidence as
\begin{equation}
\mathcal{E}(q)
=
\left(
\mathbf{x}_{r,d}^{(m)},
\mathcal{H}_{r,d},
\mathcal{M}_{r,d}
\right),
\end{equation}
where $\mathcal{H}_{r,d}$ contains supplementary epidemiological information and $\mathcal{M}_{r,d}$ records the source, indicator definition, reporting frequency, retrieval time, and most recent reporting period. When no structured historical observations are available, $\mathbf{x}_{r,d}^{(m)}=\varnothing$. The APDC uses the returned evidence to select the absent, sparse, or abundant evidence branch for subsequent analysis.

\subsection{Model Analysis and Risk Determination}
\label{sec:MARD}

The Model Analysis and Risk Determination (MARD) module converts the epidemiological evidence returned by DAP into an evidence-adaptive analysis and a corresponding risk level. Let $\mathbf{x}_{r,d}$ denote the available historical series and let $T_{\min}$ denote the minimum number of valid observations required for retrospective model evaluation (as shown in the following paragraph) and numerical forecasting. MARD distinguishes among three evidence regimes:
\begin{equation}
z=
\begin{cases}
\mathrm{absent},    & \mathbf{x}_{r,d}=\varnothing,\\
\mathrm{sparse},    & 0<T<T_{\min},\\
\mathrm{abundant},  & T\geq T_{\min}.
\end{cases}
\end{equation}
For absent and sparse evidence, MARD performs LLM-based evidence assessment. When historical evidence is abundant, it selects and applies a forecasting model according to its retrospective performance.

\paragraph{LLM-based evidence assessment.}
When historical observations are absent, MARD uses the LLM to synthesize the authoritative epidemiological information. The assessment identifies reported outbreak activity, changes in case occurrence, geographic spread, and other risk signals supported by the retrieved sources. Under sparse evidence, MARD additionally incorporates the limited observations in $\mathbf{x}_{r,d}$ as contextual information, while avoiding numerical forecasting. In both regimes, the LLM produces an evidence-grounded assessment and retains the supporting information used to determine the risk level.

\paragraph{Forecasting model pool.}
When historical evidence is abundant, MARD performs numerical forecasting using a pool $\mathcal{M}$ containing more than 30 forecasting models. As summarized in Table~\ref{tab:forecasting-models}, the pool includes statistical and rule-based methods, machine-learning models, deep neural models, and pretrained time-series foundation models. These model families capture different forms of temporal dependence, trend, seasonality, and nonlinear variation, allowing MARD to select a forecasting approach according to the characteristics of the current surveillance series.

\begin{table}[t]
\centering
\small
\caption{Forecasting models included in the numerical forecasting component of \systemname. The model pool contains 30+ statistical, machine learning, deep learning, and pretrained time-series foundation models.}
\label{tab:forecasting-models}
\begin{tabular}{p{0.28\linewidth}p{0.65\linewidth}}
\toprule
\textbf{Model family} & \textbf{Forecasting models} \\
\midrule
Statistical and rule-based & Persistence, Historical mean, Drift, Seasonal naive, Moving average variants, Exponential smoothing variants, Theta, ARIMA~\cite{box2015timeseries}, SARIMA~\cite{hyndman2008automatic}, and SARIMAX~\cite{hyndman2021forecasting}. \\
\hline
Machine learning & Linear regression, Ridge regression, Random forest, Gradient boosting, Support vector regression~\cite{smola2004svr}, XGBoost~\cite{chen2016xgboost}, and LightGBM~\cite{ke2017lightgbm}. \\
\hline
Deep Learning & LSTM~\cite{hochreiter1997lstm}, GRU~\cite{cho2014gru}, TCN~\cite{bai2018tcn}, Transformer~\cite{vaswani2017attention}, N-BEATS~\cite{oreshkin2020nbeats}, N-HiTS~\cite{challu2023nhits}, and PatchTST~\cite{nie2023patchtst}. \\
\hline
Time-series foundation models & Chronos-2~\cite{ansari2025chronos2}, TimesFM~\cite{das2024timesfm}, Moirai~2.0~\cite{liu2025moirai2}, Sundial~\cite{liu2025sundial}, Toto~2.0~\cite{khwaja2026toto2}, and Time-MoE~\cite{shi2025timemoe}. \\
\bottomrule
\end{tabular}
\end{table}

Given a historical series $\mathbf{x}=(x^1,x^2,\ldots,x^T)$ and forecast horizon $H$, each model $m\in\mathcal{M}$ produces
\begin{equation}
\hat{\mathbf{y}}^{m}_{T+1:T+H}
=
\left(
\hat{y}^{m}_{T+1},
\hat{y}^{m}_{T+2},
\ldots,
\hat{y}^{m}_{T+H}
\right).
\end{equation}
All models receive the same historical observations and predict the same future interval, enabling retrospective performance comparison under a common forecasting setting.

\paragraph{Retrospective model selection.}
Since the observations over the future interval $[T+1,T+H]$ are unavailable at prediction time, MARD evaluates each candidate model on the most recent fully observed interval. Specifically, the final $H$ observations are used as the validation target:
\begin{equation}
\mathbf{y}^{\mathrm{val}}
=
\left(
x^{T-H+1},
\ldots,
x^{T}
\right).
\end{equation}
Each model receives the preceding observations $\mathbf{x}_{1:T-H}$ and produces a retrospective prediction $ \hat{\mathbf{y}}^{m,\mathrm{val}} = \left(\hat{x}^{m}_{T-H+1},\ldots, \hat{x}^{m}_{T} \right)$.
Its retrospective error is measured using the mean squared error $ E_m = \frac{1}{H} \sum_{h=1}^{H} \left(x^{T-H+h}-\hat{x}^{m}_{T-H+h}\right)^2$.
MARD selects the model with the lowest validation error:
\begin{equation}
m^{\star}
=
\arg\min_{m\in\mathcal{M}}
E_m,
\end{equation}
The selected model is then applied to the complete series $\mathbf{x}_{1:T}$ to generate the final forecast $\hat{\mathbf{y}}^{m^{\star}}_{T+1:T+H}$.


\paragraph{Risk-level determination.}
MARD maps the available analytical evidence to one of four risk levels:
\texttt{OK}, \texttt{WATCH}, \texttt{WARNING}, or \texttt{CRITICAL}. For 
evidence, let $p_{50}$, $p_{75}$, and $p_{90}$ denote the corresponding percentiles of the historical observations, let $x_{\max}$ denote the historical maximum, and define $\hat{y}_{\max} = \max_{1\leq h\leq H} \hat{y}^{m^{\star}}_{T+h}$.
The risk level is determined from the highest to the lowest severity:
\begin{equation}
\operatorname{Risk}=
\begin{cases}
\texttt{CRITICAL}, & x^{T}\geq p_{90} \ \lor\ \hat{y}_{\max}>x_{\max}, \\
\texttt{WARNING}, & x^{T}\geq p_{75} \ \lor\ \hat{y}_{\max}\geq p_{90}, \\
\texttt{WATCH}, & x^{T}\geq p_{50} \ \lor\ \operatorname{Trend} \left( \hat{\mathbf{y}}^{m^{\star}} \right) =\mathrm{upward}, \\
\texttt{OK}, & \text{otherwise}.
\end{cases}
\end{equation}
For absent and sparse evidence, MARD retains the same four risk labels but determines the level through LLM-based reasoning over the available observations and authoritative epidemiological information. MARD returns the resulting risk level together with the selected forecast or supporting epidemiological evidence, enabling RG to present both the warning result and the basis for its determination.

\subsection{Report Generation}

The analytical branches coordinated by the APDC produce different outputs. An abundant-evidence analysis may include historical observations, temporal trends, and model forecasts, whereas absent- and sparse-evidence analyses rely more heavily on authoritative epidemiological information. Directly rendering these diverse outputs as free-form text makes it difficult to enforce a consistent report structure or verify whether each conclusion is adequately supported. The Report Generation (RG) component therefore converts the analytical outputs into a validated structured representation before generating the final report.

\paragraph{Evidence-adaptive report representation.}
RG maintains three report templates corresponding to the evidence-availability regimes: $ \mathcal{T} = \left\{ \mathcal{T}_{\mathrm{absent}}, \mathcal{T}_{\mathrm{sparse}},
\mathcal{T}_{\mathrm{abundant}} \right\}$.
Given the regime $z\in\{\mathrm{absent},\mathrm{sparse},\mathrm{abundant}\}$ and the analytical output $\mathcal{O}_z$, RG selects the corresponding template and constructs a structured report payload.
All templates contain common fields for the disease, geographic region, report date, risk level, summary, risk factors, recommended actions, and evidence sources. The absent-evidence template records the retrieved epidemiological information and the evidence supporting the warning decision. The sparse-evidence template additionally includes the available observations and their temporal context. The abundant-evidence template further records the historical trend, selected forecasting model, and future predictions. These regime-specific fields prevent unsupported forecasting results from appearing in reports without sufficient historical evidence.

\paragraph{Report validation and revision.}
Before rendering, RG validates the structured payload at the schema, value, and content levels.
The schema validator verifies the presence and type of all fields required by $\mathcal{T}_z$. The value validator checks fields with constrained values, including whether the risk level belongs to $ \mathcal{W} = \left\{ \texttt{OK}, \texttt{WATCH}, \texttt{WARNING}, \texttt{CRITICAL} \right\}$.
The content validator examines whether the report is consistent with the selected evidence regime and its supporting analytical outputs. For example, an abundant-evidence report must contain the historical series and forecasting result used in risk determination, whereas an absent-evidence report must not present an unsupported numerical forecast. If validation fails, RG returns a structured error identifying the invalid field and violated constraint. The APDC can then invoke the relevant tool to revise the analytical output or report field.

\paragraph{Unified report rendering.}
Once the payload passes all validation stages, RG invokes the rendering engine that generates the Web and PDF reports. The renderer also creates visualizations supported by the available evidence, such as historical surveillance curves, future forecasts, prediction intervals, and risk-level annotations.

\subsection{Self-improvement Mechanism}

The {self-improvement mechanism} preserves effective execution procedures from completed early-warning tasks and incorporates them into the SOP repository. A successful task records not only its final report, but also the data sources, preprocessing operations, tool and model configurations, evidence used for risk assessment, and report-generation steps followed during execution. When the task is completed and its report passes validation, these process-level records are retained as a successful execution trajectory.

The mechanism uses each successful trajectory to create or refine the SOP associated with its disease--region pair. If the task is executed through the zero-shot workflow and no corresponding SOP previously existed, the validated sequence of operations is consolidated into a new SOP. If an SOP is retrieved for the task, the successful trajectory updates its data routes, processing steps, tool configurations, or analytical procedures. The resulting SOP is written back to the SOP set $\mathcal{G}$ maintained by \systemname.

During subsequent requests, the UIL can retrieve the newly created or updated SOP, and the APDC can use it to construct the corresponding execution plan. In this way, procedures successfully applied in previous tasks become reusable system knowledge, reducing the need to reconstruct the same workflow in future executions. \systemname\ therefore improves itself through continual SOP maintenance and reuse rather than parameter updates to the underlying models.

\section{Experiments}\label{sec:experiments}

\paragraph{Dataset and Setting.}
We introduce two infectious disease forecasting datasets, China-Infectious Disease and US-Infectious Disease, using official surveillance data released by the China CDC and the U.S. CDC, respectively. The China-Infectious Disease dataset comprises five time series of respiratory pathogen positivity rates reported by the China CDC, while the US-Infectious Disease dataset contains reported case-count time series for ten nationally notifiable infectious diseases published by the U.S. CDC. These time series are reported at 1-week intervals. For each dataset, we split the time series chronologically into training, validation, and test sets at a 6:2:2 ratio. Given historical observations, the forecasting task is to predict the positivity rate or case count for the subsequent four weeks. We evaluate forecasting performance using the Mean Absolute Scaled Error (MASE) and Root Mean Squared Error (RMSE).

\paragraph{Baselines.}
We compare \systemname\ with the five  end-to-end forecasting systems, including AutoGluon-TimeSeries~\cite{shchur2023autogluon}, the epidemic-specific pretrained forecaster CAPE~\cite{liu2026pre}, the TimeSeriesScientist agent~\cite{zhao2025timeseriesscientist}, the epi4cast weighted ensemble~\cite{wang2026naive}, and the fine-tuned LLM forecasting system Predict NIDs~\cite{wu2026large}.
We additionally construct two general-purpose LLM baselines by combining API models, DeepSeek V4 Flash and GPT-4o mini.

\subsection{Main Results}
Table~\ref{tab:main-system-results} reports each model's average forecasting performance on the China-Infectious Disease and US-Infectious Disease datasets. Our model achieves the best performance across all evaluation metrics on both datasets, outperforming the second-best model by 9.68\% and 14.04\%, respectively. Moreover, compared with directly using DeepSeek-V4 for forecasting, \systemname\ achieves a 23.03\% performance improvement on the US-Infectious Disease dataset, demonstrating the effectiveness of our agent planning and decision center in coordinating and invoking different components to accomplish the forecasting task.

\begin{table}[t]
\centering
\small
\caption{Performance comparison on the China and US infectious disease datasets. The lower is the better.}
\label{tab:main-system-results}
\resizebox{0.8\linewidth}{!}{%
\begin{tabular}{lrrrr}
\toprule
& \multicolumn{2}{c}{China-Infectious Disease} & \multicolumn{2}{c}{US-Infectious Disease}\\
\cmidrule(lr){2-3}\cmidrule(lr){4-5}
System & MASE $\downarrow$ & RMSE $\downarrow$ & MASE $\downarrow$ & RMSE $\downarrow$\\
\midrule
LLM Predict NIDs & $4.26 \pm 0.68$ & $4.12 \pm 0.97$ & $2.27 \pm 0.64$ & $64.91 \pm 13.26$\\
epi4cast Weighted Ensemble & $2.05 \pm 0.08$ & $2.02 \pm 0.14$ & $1.37 \pm 0.14$ & $40.46 \pm 2.26$\\
TimeSeriesScientist & $2.02 \pm 0.04$ & $1.95 \pm 0.09$ & $1.20 \pm 0.04$ & $40.30 \pm 1.12$\\
AutoGluon-TimeSeries & $1.76 \pm 0.03$ & $1.75 \pm 0.08$ & $1.09 \pm 0.06$ & $34.08 \pm 1.35$\\
CAPE & $1.78 \pm 0.00$ & $1.66 \pm 0.00$ & $1.18 \pm 0.00$ & $36.55 \pm 0.00$\\
DeepSeek V4 Flash ReAct & $1.54 \pm 0.00$ & $1.56 \pm 0.00$ & $1.26 \pm 0.00$ & $36.90 \pm 0.00$\\
GPT-4o mini ReAct & $1.56 \pm 0.00$ & $1.60 \pm 0.00$ & $1.36 \pm 0.00$ & $45.81 \pm 0.00$\\

\textbf{\systemname} & $\mathbf{1.38 \pm 0.01}$ & $\mathbf{1.42 \pm 0.01}$ & $\mathbf{0.90 \pm 0.01}$ & $\mathbf{30.45 \pm 0.78}$\\
\bottomrule
\end{tabular}%
}
\end{table}

\subsection{Analysis of Time-Series Foundation Models}

\begin{table}[t]
\centering
\small
\caption{Comparison of forecasting performance using the fixed time-series foundation model.}
\label{tab:tsfm-ablation}
\resizebox{0.7\linewidth}{!}{%
\begin{tabular}{lrrrr}
\toprule
& \multicolumn{2}{c}{China-Infectious Disease} & \multicolumn{2}{c}{US-Infectious Disease}\\
\cmidrule(lr){2-3}\cmidrule(lr){4-5}
Internal TSFM & MASE $\downarrow$ & RMSE $\downarrow$ & MASE $\downarrow$ & RMSE $\downarrow$\\
\midrule
Sundial & $1.71 \pm 0.01$ & $1.73 \pm 0.00$ & $1.06 \pm 0.00$ & $32.56 \pm 0.28$\\
Time-MoE & $1.52 \pm 0.00$ & $1.55 \pm 0.00$ & $1.08 \pm 0.00$ & $33.14 \pm 0.00$\\
TimesFM 2.5 & $1.50 \pm 0.00$ & $1.51 \pm 0.00$ & $0.99 \pm 0.00$ & $31.19 \pm 0.00$\\
Moirai 2.0 & $1.62 \pm 0.00$ & $1.80 \pm 0.00$ & $0.92 \pm 0.00$ & $29.72 \pm 0.00$\\
Chronos-2 & $1.41 \pm 0.00$ & $1.48 \pm 0.00$ & $0.98 \pm 0.00$ & $32.09 \pm 0.00$\\
Toto 2.0 & $1.34 \pm 0.00$ & $1.36 \pm 0.00$ & $0.94 \pm 0.00$ & $30.51 \pm 0.00$\\
\bottomrule
\end{tabular}
}
\end{table}
We evaluate the effectiveness of the forecasting model pool by comparing \systemname\ with variants that use a single fixed time-series foundation model for forecasting. The results are presented in Table~\ref{tab:tsfm-ablation}. We observe that Moirai 2.0, Chronos-2, and Toto 2.0 consistently rank among the top three models on both the China and U.S. infectious disease datasets. However, the best-performing model varies across datasets: Toto 2.0 achieves the best performance in China, whereas Moirai 2.0 performs best in the United States. These results indicate that no single time-series foundation model consistently performs best across different datasets. This demonstrates the effectiveness of using a forecasting model pool to select the most suitable model for different forecasting scenarios. 



\subsection{Case Study}

\begin{figure}[htbp]
\centering
\begin{minipage}[t]{0.49\linewidth}
\centering
\includegraphics[width=\linewidth]{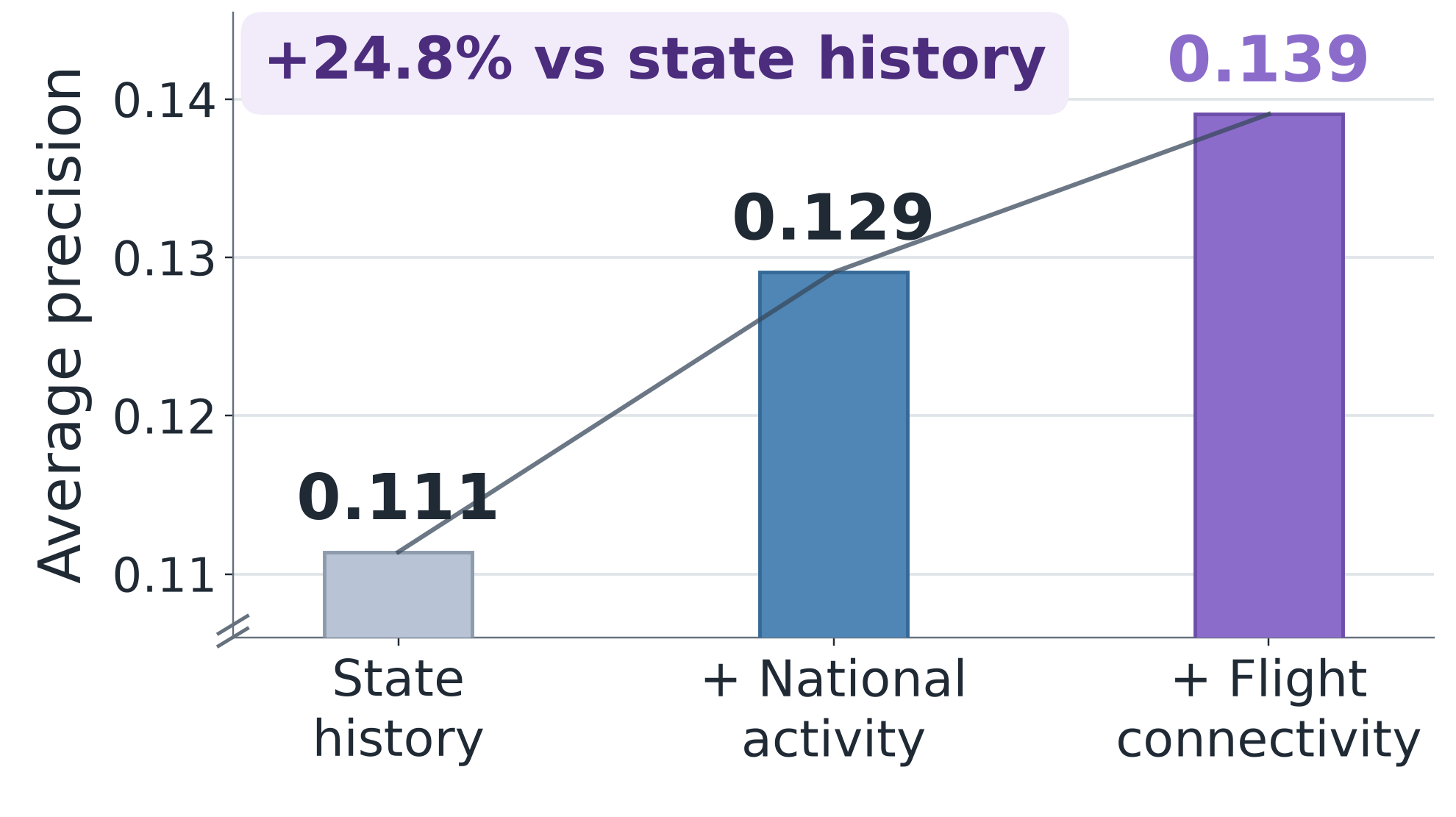}\\[-0.5ex]
\small\textbf{(a)} Incremental predictive value
\end{minipage}\hfill
\begin{minipage}[t]{0.49\linewidth}
\centering
\includegraphics[width=\linewidth]{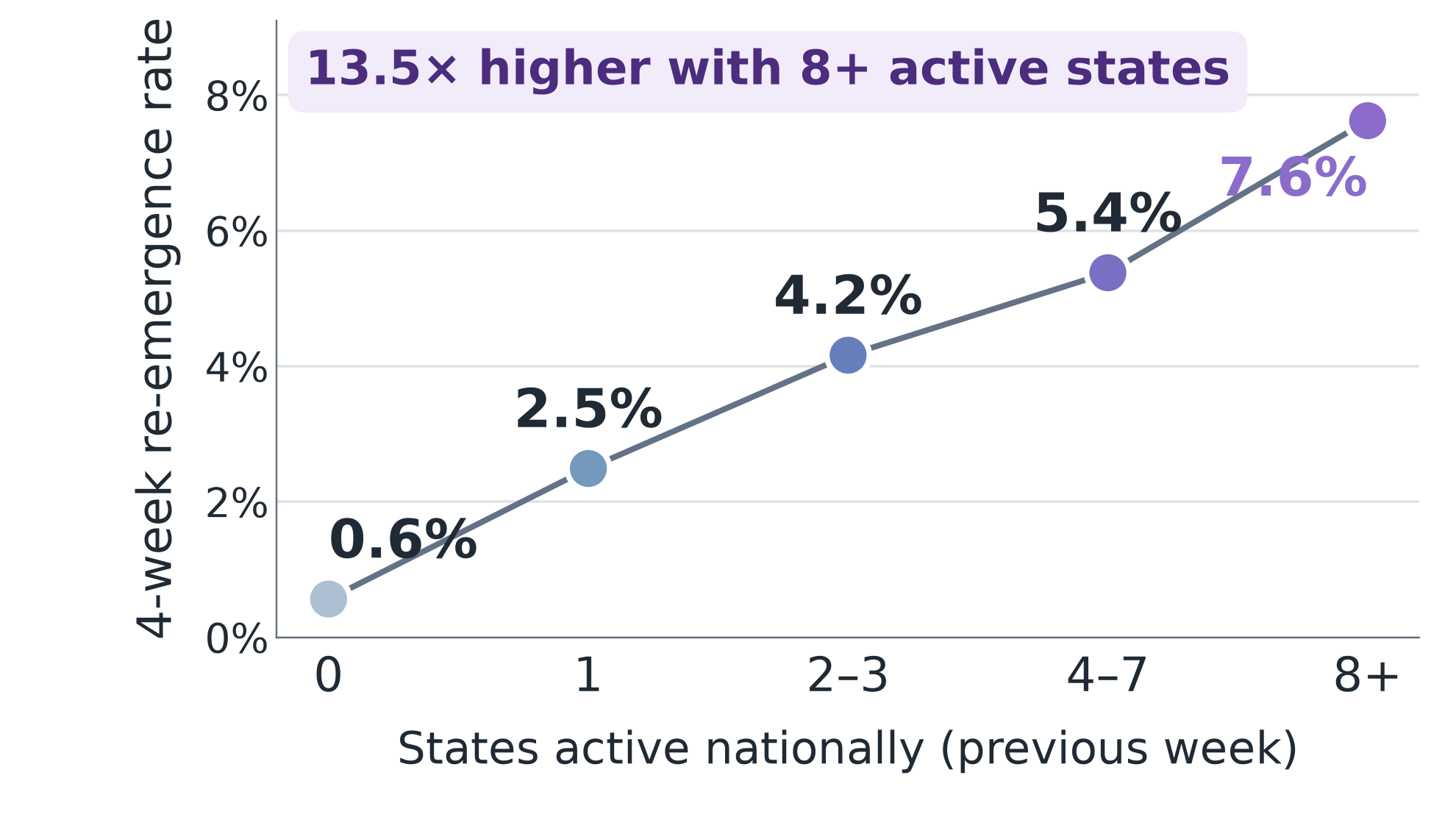}\\[-0.5ex]
\small\textbf{(b)} National activity gradient
\end{minipage}
\caption{Multi-source evidence for four-week state-level reported-case re-emergence. (a) Average precision as national activity and flight-connectivity features are added to state history. (b) Observed re-emergence rates stratified by the number of states reporting the same disease in the previous week.}
\label{fig:multisource-risk}
\end{figure}

\paragraph{Multi-source re-emergence signals.} We further investigate the effectiveness of incorporating additional multi-source information on the U.S. infectious disease dataset. Specifically, we focus on states that have reported no cases for several consecutive weeks. We then examine whether national disease activity and interstate connectivity provide useful information for predicting case re-emergence.

As shown in Figure~\ref{fig:multisource-risk}(a), incorporating national disease activity improves forecasting performance by 16.22\%. This suggests that a state's disease case is more likely to re-emerge when disease activity is more widespread across other states. Adding interstate flight connectivity further improves performance, resulting in a 24.8\% improvement over using only state-level historical information. This indicates that interstate connectivity provides information for identifying states that are more exposed to external disease activity.

Figure~\ref{fig:multisource-risk}(b) shows a consistent pattern. The likelihood of case re-emergence increases markedly as disease activity becomes more widespread nationwide. These findings suggest that re-emergence risk depends not only on a state's recent disease history, but also on broader patterns of disease activity and interstate connectivity, demonstrating the value of multi-source information for improving early warning of case re-emergence.

\begin{figure}[t]
\centering
\begin{minipage}[t]{0.49\linewidth}
\centering
\includegraphics[width=\linewidth]{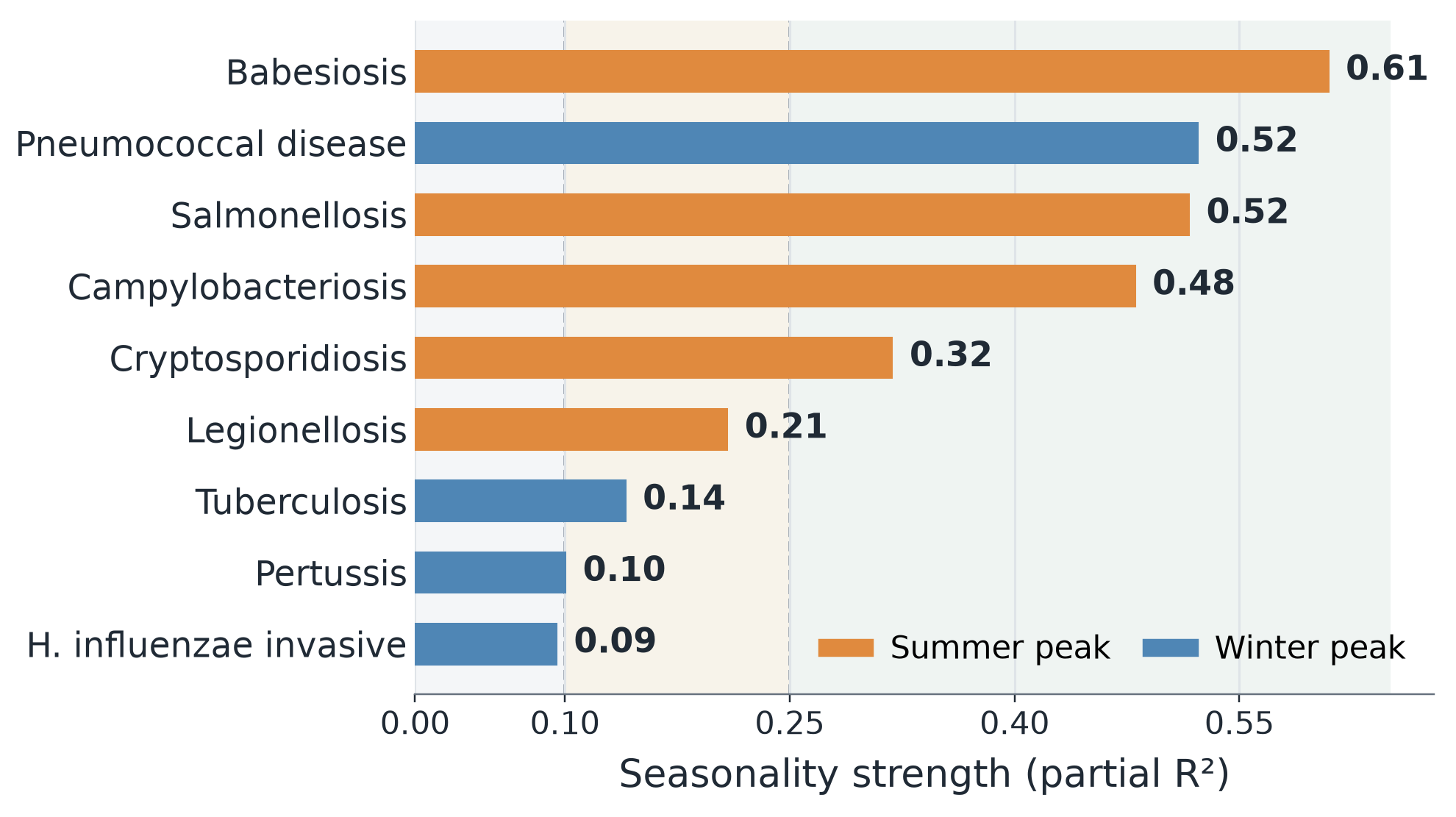}\\[-0.5ex]
\small\textbf{(a)} Disease seasonality
\end{minipage}\hfill
\begin{minipage}[t]{0.49\linewidth}
\centering
\includegraphics[width=\linewidth]{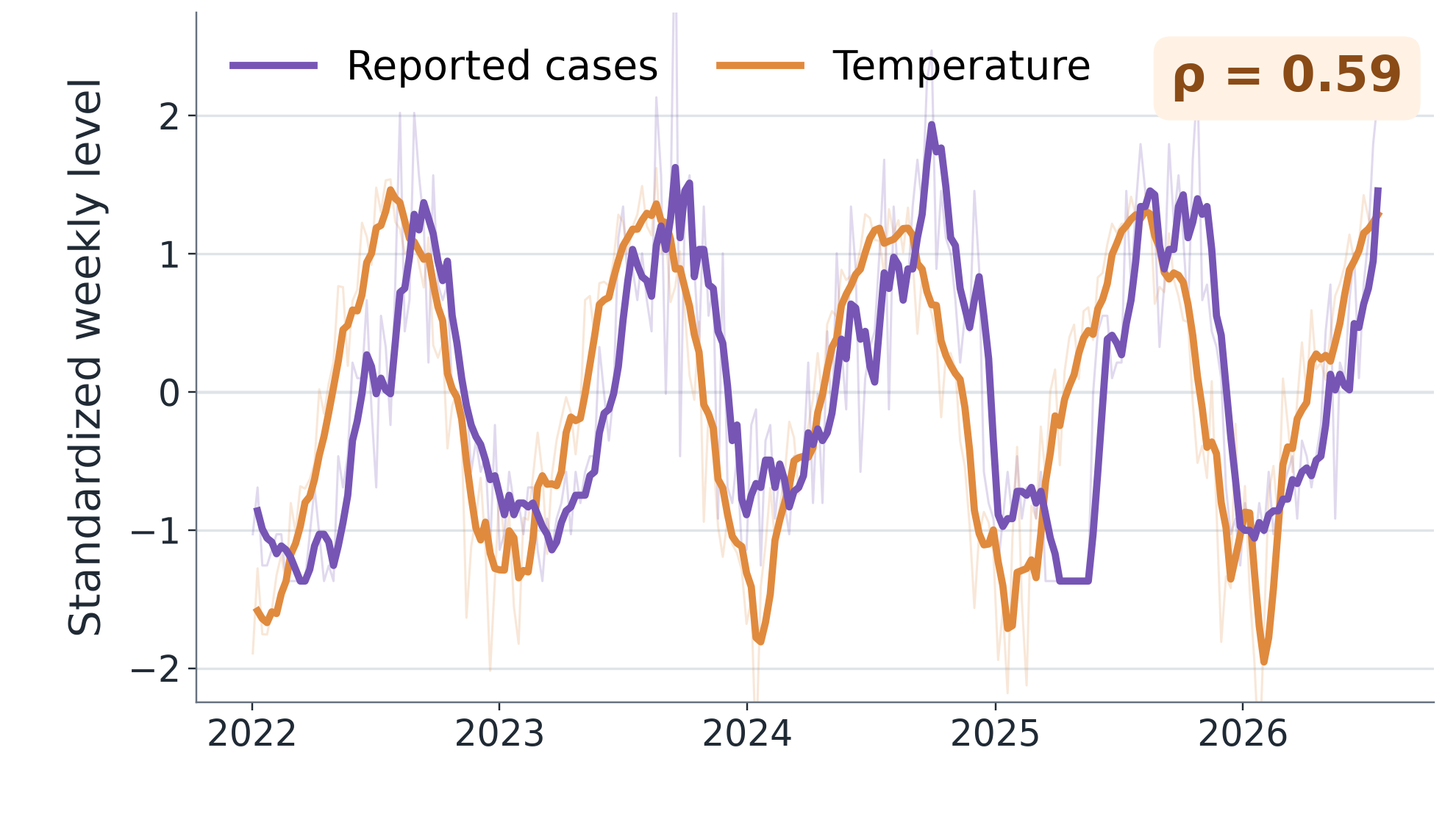}\\[-0.5ex]
\small\textbf{(b)} Temperature alignment
\end{minipage}

\medskip
\begin{minipage}[t]{0.49\linewidth}
\centering
\includegraphics[width=\linewidth]{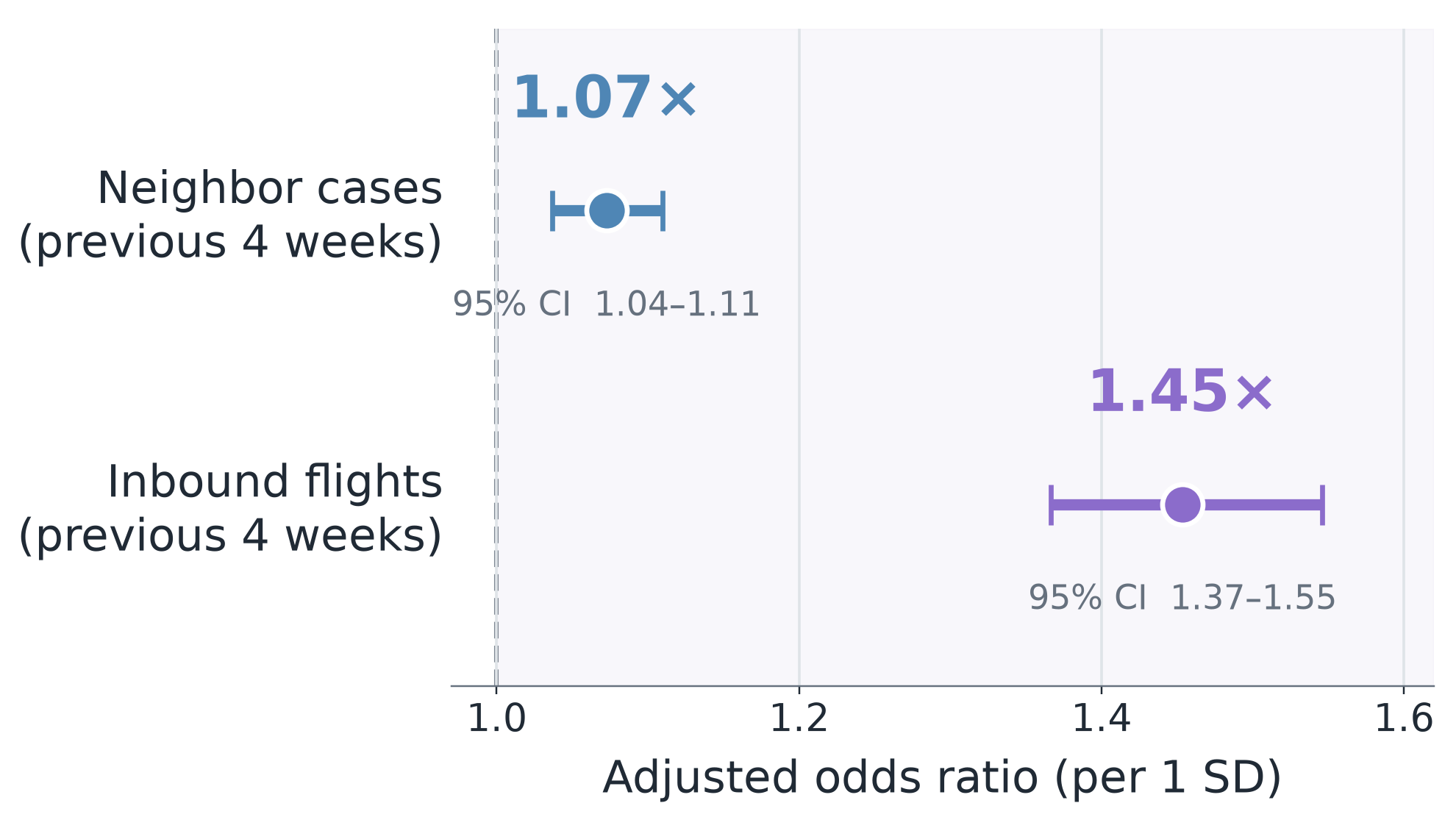}\\[-0.5ex]
\small\textbf{(c)} Adjusted spatial associations
\end{minipage}\hfill
\begin{minipage}[t]{0.49\linewidth}
\centering
\includegraphics[width=\linewidth]{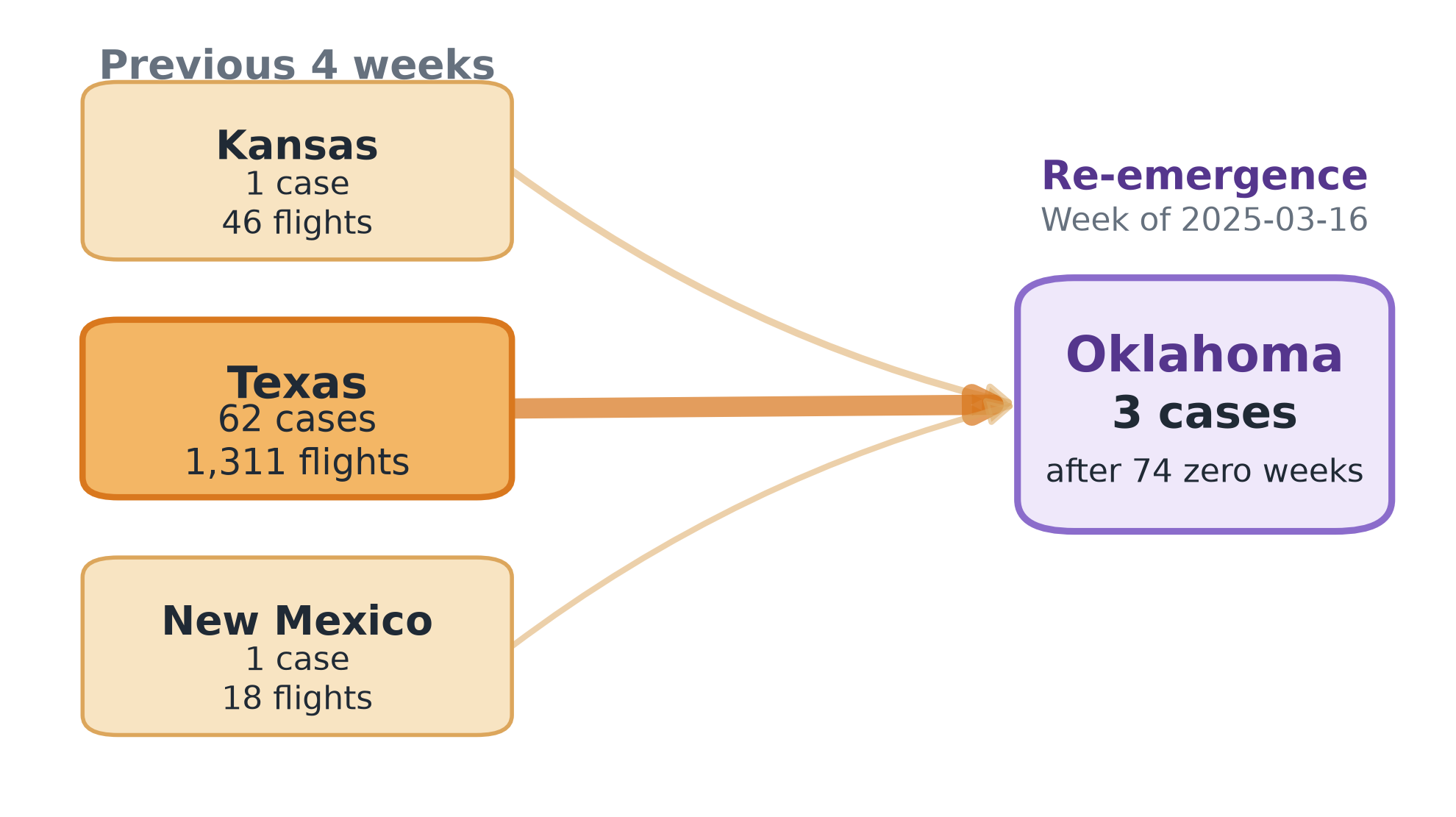}\\[-0.5ex]
\small\textbf{(d)} Oklahoma measles example
\end{minipage}
\caption{Temporal and geographic context for state-level surveillance. (a) Annual-harmonic partial $R^2$, colored by peak season. (b) Standardized weekly Arkansas salmonellosis reports and temperature. (c) Adjusted ORs per one standard deviation from disease--week matched analysis. (d) Neighboring cases and inbound flights during the four weeks preceding reported measles re-emergence in Oklahoma. These panels describe predictive associations rather than causal transmission pathways.}
\label{fig:context-signals}
\end{figure}

\paragraph{Seasonal, meteorological, spatial evidence.}
Figure~\ref {fig:context-signals} illustrates the relationships between disease cases and temporal and geographic information. We observe seasonal patterns in disease incidence, with substantial variation across different diseases. Meteorological conditions also exhibit temporal patterns that are closely associated with local disease dynamics. These findings suggest that disease activity in neighboring regions and interstate connectivity can add context to disease forecasting.

Figure~\ref{fig:context-signals}(d) presents a measles case study in Oklahoma. Oklahoma had reported no measles cases for 74 consecutive weeks. However, three states with strong flight connectivity to Oklahoma reported different levels of measles activity during the preceding four weeks. Oklahoma subsequently reported measles cases. This example shows that signals associated with case re-emergence can be traced to specific information from different data sources, such as recent disease activity in connected states and interstate mobility patterns. Such traceable evidence helps explain the contextual factors behind the model's forecasts.

\paragraph{API-token cost.}
In the early warning process, inference cost is also an important factor in assessing the system's practical applicability. Because different large language models incur different token consumption when performing risk prediction, we compare \systemname's inference costs with DeepSeek V4 and GPT-4o. As shown in Table~\ref{tab:api-efficiency}, \systemname reduces API calls for risk prediction by 27.27\% while reducing average token consumption per call by 32.89\%. These results show that \systemname can predict risk with fewer API calls and lower token overhead, improving computational efficiency and reducing deployment costs while maintaining strong early-warning performance.

\begin{table}[htbp]
\centering
\small
\caption{Forecast accuracy and full-workflow external LLM-token usage on two datasets. Costs include forecasting, planning, orchestration, and report assembly. Lower is better.}
\label{tab:api-efficiency}
\resizebox{0.9\linewidth}{!}{%
\begin{tabular}{lrrrrr}
\toprule
System & China MASE $\downarrow$ & US MASE $\downarrow$ & Total calls $\downarrow$ & Total tokens $\downarrow$ & Tokens/task $\downarrow$\\
\midrule
GPT-4o mini ReAct & $1.559^{\dagger}$ & $1.357^{\dagger}$ & $\approx 495^{\ddagger}$ & $\approx 1{,}035{,}455^{\ddagger}$ & $\approx 23{,}010^{\ddagger}$\\
DeepSeek V4 Flash ReAct & $1.545^{\dagger}$ & $1.261^{\dagger}$ & $\approx 495^{\ddagger}$ & $\approx 1{,}042{,}159^{\ddagger}$ & $\approx 23{,}159^{\ddagger}$\\
\textbf{EpiGuard} & $\mathbf{1.377 \pm 0.009}$ & $\mathbf{0.903 \pm 0.014}$ & \textbf{$\approx 360^{\ddagger}$} & \textbf{$\approx 694{,}905^{\ddagger}$} & \textbf{$\approx 15{,}442^{\ddagger}$}\\
\bottomrule
\end{tabular}%
}
\par\vspace{2pt}
\begin{minipage}{0.9\linewidth}
\footnotesize $\dagger$ denotes one API execution. $\ddagger$ denotes full-workflow token accounting based on the representative complete report trace; local computation is excluded.
\end{minipage}
\end{table}

\begin{figure}[htbp]
\centering
\begin{minipage}[t]{0.325\linewidth}
\centering
\includegraphics[width=\linewidth]{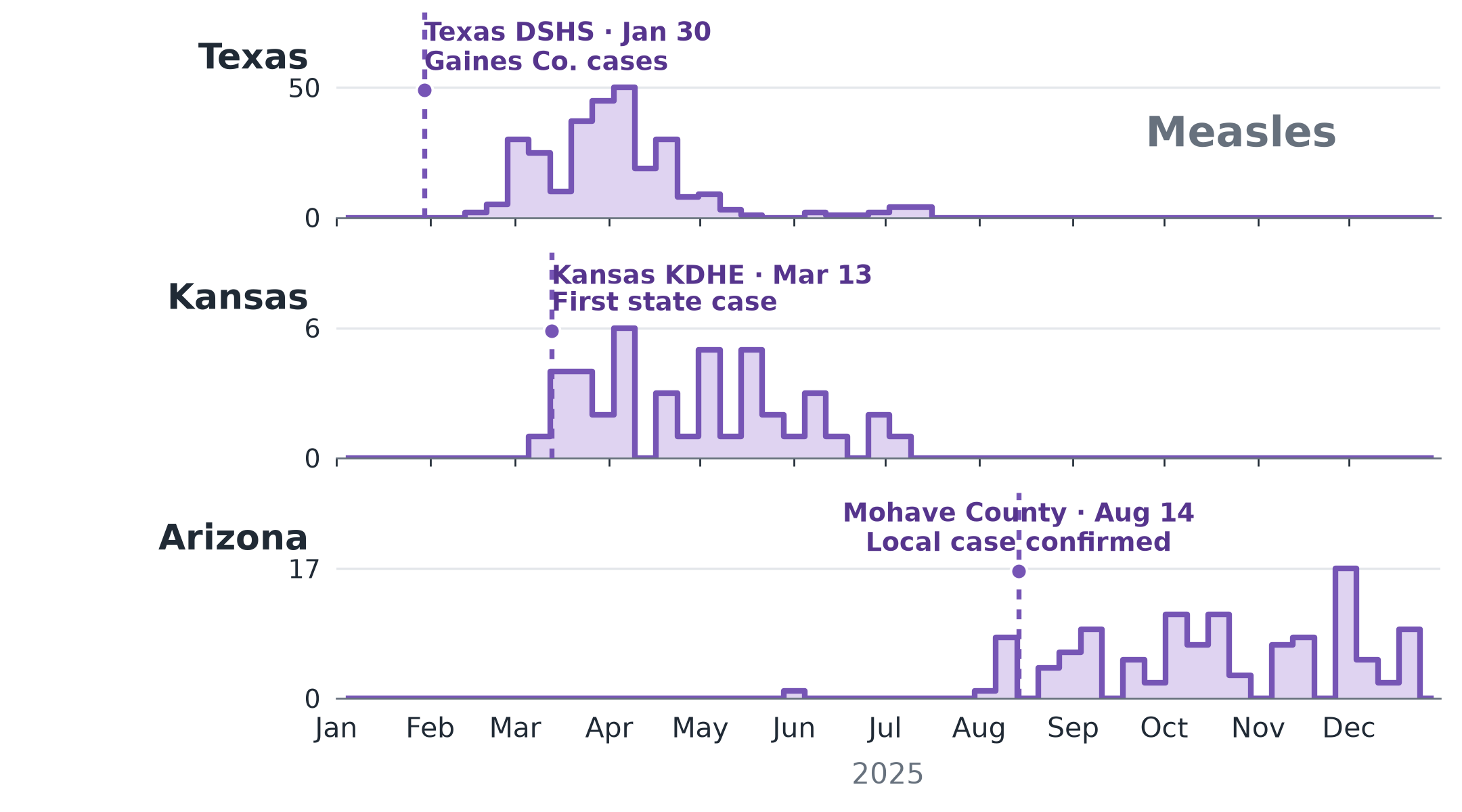}\\[-0.5ex]
\small\textbf{(a)} Measles
\end{minipage}\hfill
\begin{minipage}[t]{0.325\linewidth}
\centering
\includegraphics[width=\linewidth]{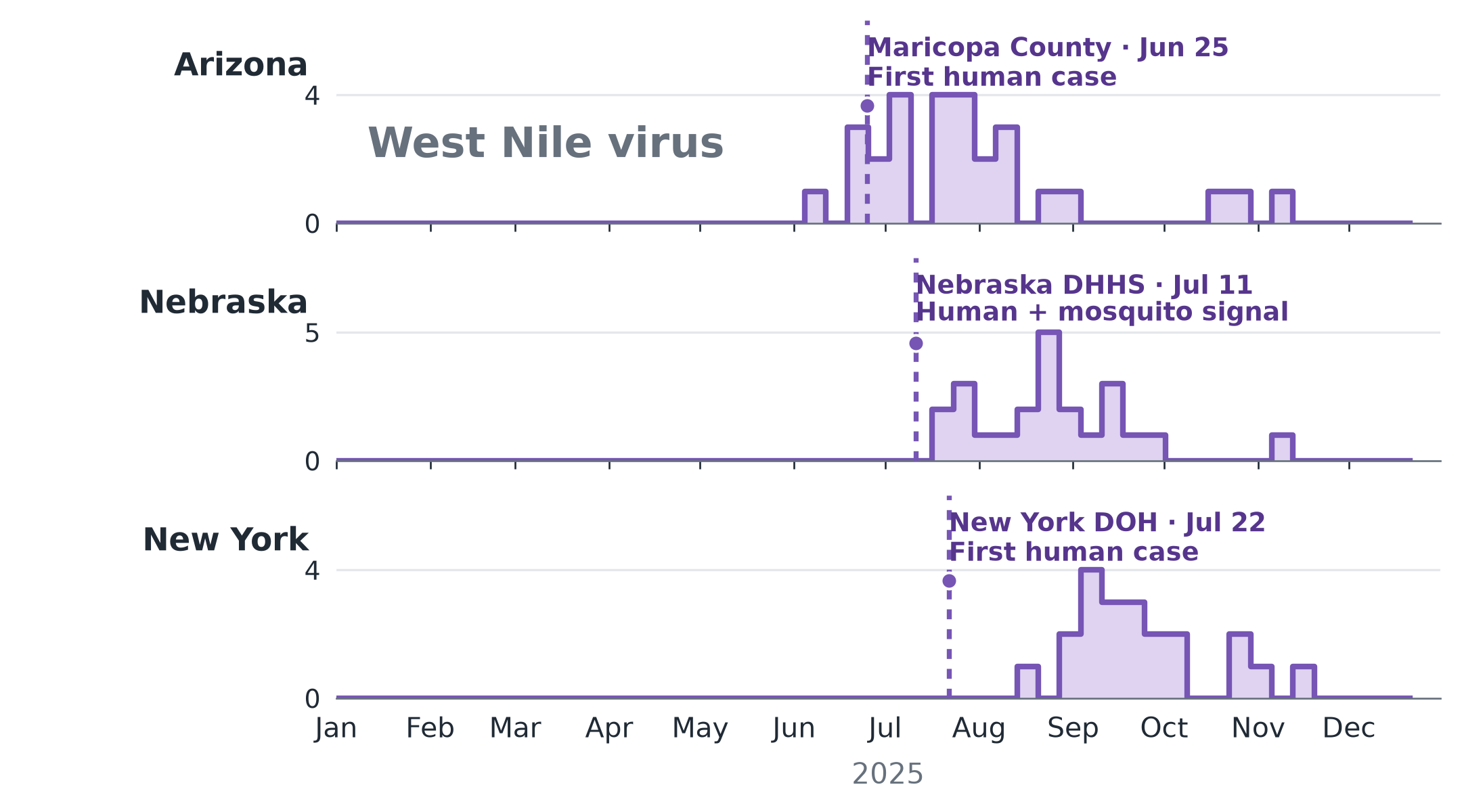}\\[-0.5ex]
\small\textbf{(b)} West Nile virus
\end{minipage}\hfill
\begin{minipage}[t]{0.325\linewidth}
\centering
\includegraphics[width=\linewidth]{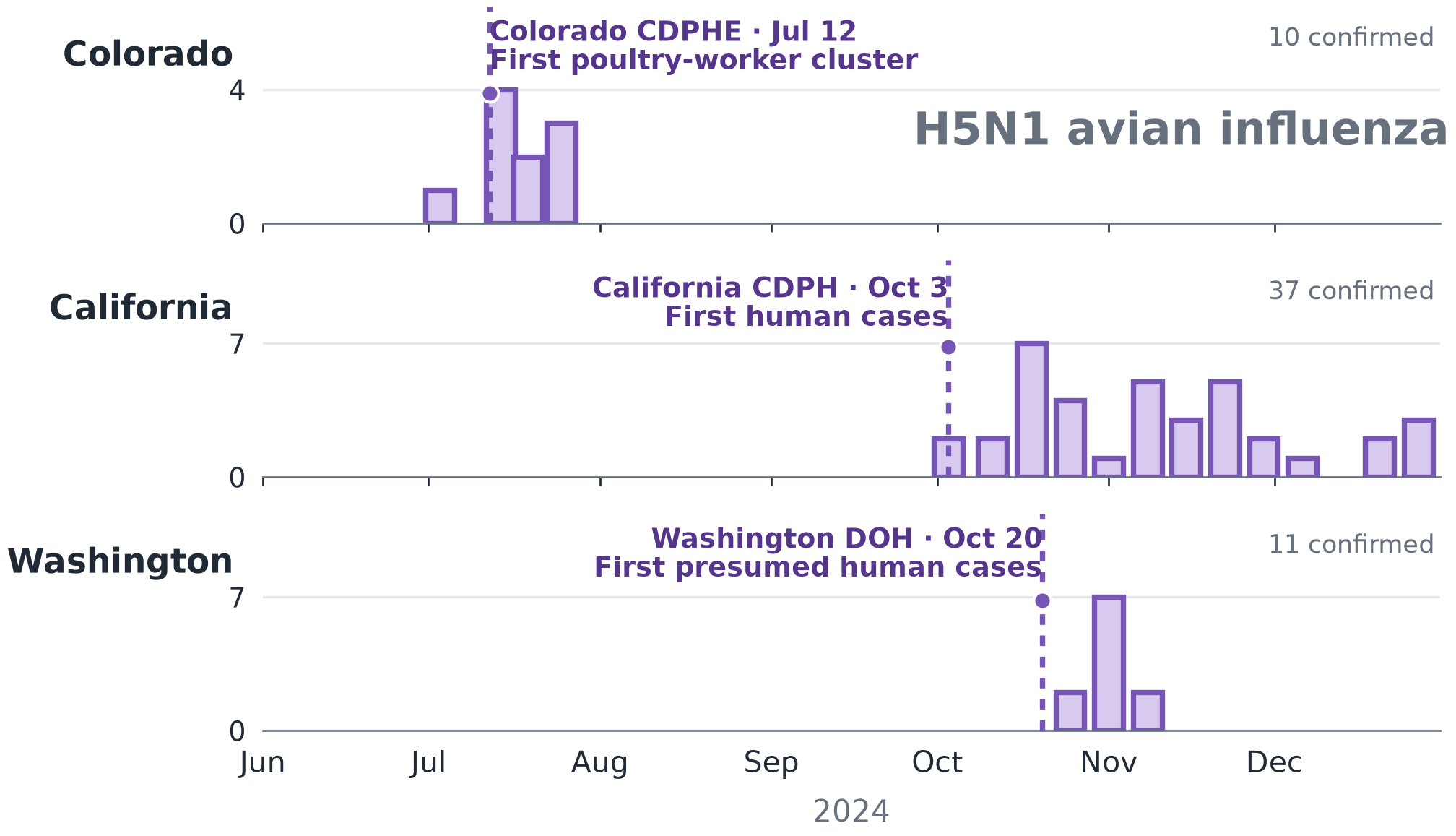}\\[-0.5ex]
\small\textbf{(c)} H5N1
\end{minipage}
\caption{Incremental information in official public-health notices. State-level case trajectories are paired with dated official-notice markers for (a) measles, (b) West Nile virus, and (c) H5N1 avian influenza. The notices provide location, exposure, and event-type context beyond case counts alone.}
\label{fig:official-notices}
\end{figure}

\FloatBarrier

\paragraph{Official-notice evidence.}
As shown in Figure~\ref{fig:official-notices}, official state and local notices provide event-level semantics that are not available from numerical case trajectories alone. Across multiple disease families, these notices tend to emerge around the onset or early development of local activity and contribute complementary information about location, exposure setting, and event type. This makes official notices particularly useful for interpreting sparse or newly emerging signals, where case counts alone may provide limited context.

\begin{wrapfigure}{r}{0.43\linewidth}
\vspace{-0.8\baselineskip}
\centering
\includegraphics[width=\linewidth]{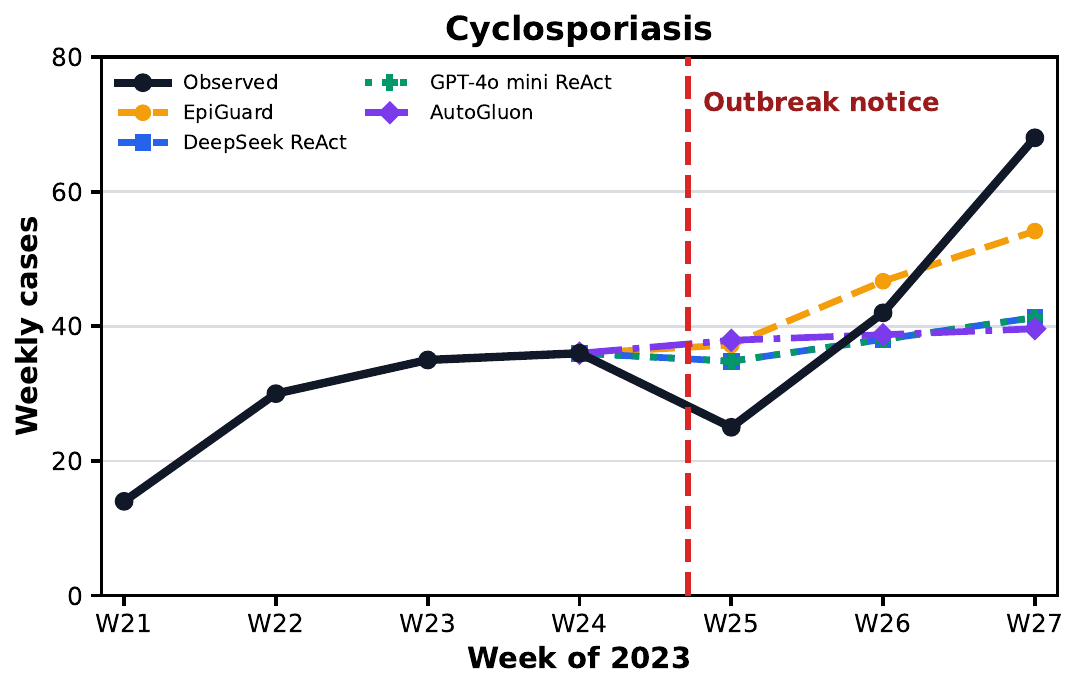}
\caption{Cyclosporiasis outbreak forecasting.}
\label{fig:cyclosporiasis-outbreak-forecast}
\vspace{-0.8\baselineskip}
\end{wrapfigure}
\paragraph{Timely outbreak anticipation.}
To evaluate the timeliness of outbreak warnings, we conduct a case study to examine whether emerging outbreaks could be accurately anticipated before an official alert is released. We use the historical 12 weeks of case counts to forecast the next four weeks. As shown in Figure~\ref{fig:cyclosporiasis-outbreak-forecast}, EpiGuard captures the upward trend before the official outbreak notice is issued and produces forecasts that closely follow the subsequent increase in reported cases. In contrast, the baseline methods largely remain near the pre-outbreak risk level and show only limited responses to the emerging change. These results suggest that EpiGuard can identify outbreak signals earlier and more accurately than the baseline methods, demonstrating its timeliness for infectious disease warning.

\section{Conclusion}
\label{sec:conclusion}

This paper presents \systemname, an autonomous and continually self-improving agentic platform for infectious-disease early warning. \systemname\ proactively maintains an up-to-date epidemiological evidence repository and employs hierarchical planning to adapt its analytical workflow across absent, sparse, and abundant evidence regimes. It combines LLM-based evidence assessment with a heterogeneous forecasting model pool, generates validated structured reports, and consolidates successful execution trajectories into reusable and continually refined SOPs.
Experiments on official data released by the China CDC and the U.S. CDC demonstrate the effectiveness of \systemname\ for infectious-disease forecasting. Compared with the second-best methods, \systemname\ achieves average relative improvements of 9.68\% and 14.04\% on the China and U.S. CDC data, respectively. Further analysis reveals that the relative performance of individual time-series foundation models varies across surveillance settings, indicating that no single model consistently performs best across different scenarios. This variability supports the design of our forecasting pool and task-adaptive model selection strategy, which dynamically leverages different models rather than relying on a single forecasting model.


\clearpage
\bibliographystyle{epiguard}
\bibliography{main}

\end{document}